\documentclass[letterpaper, 10 pt, conference]{ieeeconf}

\IEEEoverridecommandlockouts 

\usepackage[dvipsnames]{xcolor}
\definecolor{dark-green}{RGB}{12,80,12}

\usepackage{url}
\usepackage{dsfont}
\usepackage{amsmath}
\usepackage{amssymb}
\usepackage{cuted}
\usepackage{multirow} %
\usepackage{multicol}
\usepackage{balance}
\let\labelindent\relax
\usepackage[shortlabels]{enumitem}

\usepackage{tabularx}
\newcolumntype{C}[1]{>{\centering\let\newline\\\arraybackslash\hspace{0pt}}m{#1}} %
\newcolumntype{L}[1]{>{\let\newline\\\arraybackslash\hspace{0pt}}m{#1}} %

\usepackage[pdfencoding=auto, hidelinks]{hyperref}
\usepackage[mathscr]{euscript}

\usepackage[export]{adjustbox}
\newcolumntype{P}[1]{>{\centering\arraybackslash}p{#1}}

\iffalse %
  \newcommand{\todo}[1]{\noindent}
\else
  \newcommand{\todo}[1]{\textcolor{red}{\bf [Todo: #1]}}
  
\fi

\usepackage{booktabs}
\usepackage{xspace}
\usepackage{caption}
\usepackage{graphicx} 
\usepackage{color}
\usepackage{siunitx}
\usepackage{subcaption}
\usepackage[hang,flushmargin,symbol]{footmisc}

\usepackage{microtype}
\usepackage{rotating}
\newcommand{\secref}[1]{Sec.~\ref{#1}}
\renewcommand{\eqref}[1]{Eq.~(\ref{#1})}
\newcommand{\figref}[1]{Fig.~\ref{#1}}
\newcommand{\tabref}[1]{Tab.~\ref{#1}}

\usepackage{tabularx}
\usepackage{colortbl}
\usepackage{bm}
\usepackage{longtable}
\usepackage{array, makecell} %
\newcolumntype{Y}{>{\centering\arraybackslash}X}
\newcolumntype{Z}{>{\raggedleft\arraybackslash}X}

\usepackage{array}
\usepackage{multirow}
\usepackage{pifont}%
\usepackage{cite}
\usepackage[export]{adjustbox}
\usepackage[resetlabels]{multibib}
\newcites{New}{References}

\newcommand{\net}{RaLF}

\usepackage{tikz}
\usepackage{algpseudocode}
\usepackage{algorithm}
\usepackage[nolist,nohyperlinks]{acronym}
\newacro{BEV}[BEV]{Bird's-Eye-View}
\newacro{DNN}[DNN]{Deep Neural Network}
\newacro{CNN}[CNN]{Convolutional Neural Network}
\newacro{GAN}[GAN]{Generative Adversarial Network}
\newacro{RANSAC}[RANSAC]{Random Sample Consensus}
\newacro{GRU}[GRU]{Gated Recurrent Unit}
\newacro{VPR}[VPR]{Visual Place Recognition}

\iffalse %
  \newcommand{\cattaneo}[1]{\noindent}
  \newcommand{\nayak}[1]{\noindent}
  \newcommand{\abhi}[1]{\noindent}
  \newcommand{\todo}[1]{\noindent}
\else
  \newcommand{\cattaneo}[1]{\textcolor{orange}{\bf [DC: #1]}}
  \newcommand{\nayak}[1]{\textcolor{purple}{\bf [AN: #1]}}
  \newcommand{\abhi}[1]{\textcolor{green}{\bf [AV: #1]}}
\fi

\newlength{\spaceabove}
\newlength{\spacebelow}
\DeclareSIUnit{\rad}{rad}

\renewcommand{\baselinestretch}{0.99}

\title{\LARGE \bf
\net{}: Flow-based Global and Metric Radar Localization\\in LiDAR Maps%
}

\author{Abhijeet Nayak\textsuperscript{$\ast$}, Daniele Cattaneo\textsuperscript{$\ast$}, and Abhinav Valada %
\thanks{© 2024 IEEE. Personal use of this material is permitted. Permission from IEEE must be obtained for all other uses, in any current or future media, including reprinting/republishing this material for advertising or promotional purposes, creating new collective works, for resale or redistribution to servers or lists, or reuse of any copyrighted component of this work in other works.}%
\thanks{\textsuperscript{$\ast$}These authors contributed equally to this work.}%
\thanks{Department of Computer Science, University of Freiburg, Germany.}%
\thanks{This work was funded by the German Research Foundation (DFG) Emmy Noether Program grant number 468878300.}}

\begin{document}

\maketitle
\thispagestyle{empty}
\pagestyle{empty}

\begin{abstract}
Localization is paramount for autonomous robots. While camera and LiDAR-based approaches have been extensively investigated, they are affected by adverse illumination and weather conditions. Therefore, radar sensors have recently gained attention due to their intrinsic robustness to such conditions.
In this paper, we propose \net{}, a novel deep neural network-based approach for localizing radar scans in a LiDAR map of the environment, by jointly learning to address both place recognition and metric localization.
\net{} is composed of radar and LiDAR feature encoders, a place recognition head that generates global descriptors, and a metric localization head that predicts the 3-DoF transformation between the radar scan and the map.
We tackle the place recognition task by learning a shared embedding space between the two modalities via cross-modal metric learning.
Additionally, we perform metric localization by predicting pixel-level flow vectors that align the query radar scan with the LiDAR map.
We extensively evaluate our approach on multiple real-world driving datasets and show that \net{} achieves state-of-the-art performance for both place recognition and metric localization. Moreover, we demonstrate that our approach can effectively generalize to different cities and sensor setups than the ones used during training. We make the code and trained models publicly available at \url{http://ralf.cs.uni-freiburg.de}.
\end{abstract}
\section{Introduction}

Localization is pivotal for any autonomous robot, whether it operates in controlled environments such as factory floors or human-centric environments such as pedestrian zones and sidewalks. It is particularly important for the latter case, where the safety of other road users is of utmost importance. While Global Navigation Satellite Systems (GNSSs) are widely used for outdoor localization, their accuracy and reliability strongly deteriorate in urban canyons. Therefore, localization systems that rely on alternative modalities are essential for autonomous robots operating in such environments.

While vision-based localization~\cite{vodisch2023covio,ballardini2019visual, vodisch2022continual,valada2018deep} has been extensively studied, their performance suffers in adverse conditions such as night and rain. To overcome these limitations, methods based on LiDARs have been proposed~\cite{cattaneo2022lcdnet,arce2023padloc} due to their high precision and robustness to illumination changes. However, LiDARs are also affected by extreme weather conditions such as fog and snow, which can drastically reduce their range and accuracy. Moreover, their high cost and large size make them unsuitable for large-scale deployment. Due to these factors, recent work investigates exploiting radar scans for localization~\cite{yin2021radar,jang2023raplace}. Radars are intrinsically robust to both weather and illumination conditions, making them a promising alternative to cameras and LiDARs.
Most of existing radar-based localization methods~\cite{yin2021radar,jang2023raplace} compare the onboard radar measurement with a pre-built radar map of the environment. As radar maps are not readily available as of today, these methods require a first mapping phase of each environment in which the robot will be deployed. LiDAR maps, on the other hand, are becoming increasingly available, thanks to the growing demand for high-definition maps.

\begin{figure}[!t]
    \centering
    \includegraphics[width=0.92\linewidth]{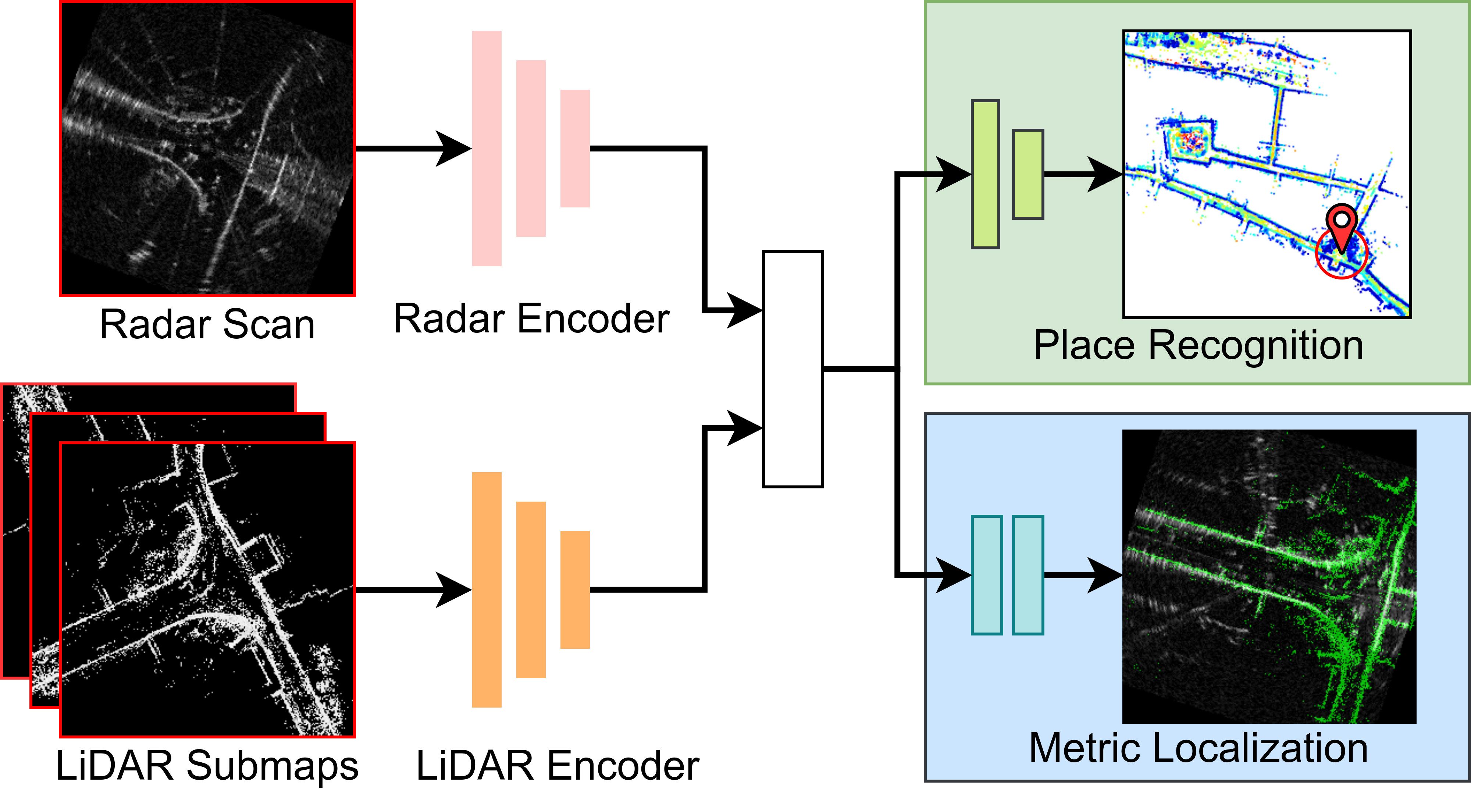}
    \caption{Our proposed \net{} localizes a radar scan within a LiDAR map both at a global (place recognition) and metric scale.}
    \label{fig:overview}
    \vspace{-13pt}
\end{figure}

A small number of works have been proposed to address the cross-modal task of localizing a radar scan within a LiDAR map, however, they only focus on a part of the localization problem, either place recognition or metric localization. On the one hand, place recognition methods~\cite{yin2021radar} provide global localization, but with inaccurate precision. On the other hand, metric localization methods~\cite{yin2021rall,yin2023radar} provide accurate localization but require an initial coarse position as input.
Therefore, for a complete localization system, a place recognition approach has to be combined with a metric localization method, introducing inefficiencies, as sensor data has to be processed by two separate approaches.

In this paper, we propose \net{}, a novel \ac{DNN}-based method for radar localization in prior LiDAR maps. Differently from existing radar-LiDAR localization methods, \net{} is, to the best of our knowledge, the first method to jointly address both place recognition and metric localization.
We reformulate the metric localization task as a flow estimation problem, where we aim at predicting pixel-level correspondences between the radar and LiDAR samples, which are subsequently used to estimate a \mbox{3-DoF} transformation.
For place recognition, we leverage a combination of same-modal and cross-modal metric learning to learn a shared embedding space where features from both modalities can be compared against each other.
We evaluate place recognition and metric localization performance of our approach on three real-world driving datasets, namely, Oxford Radar Robotcar~\cite{barnes2020oxford}, MulRan~\cite{kim2020mulran}, and Boreas~\cite{burnett2023boreas}. We compare our method against  methods, and show that \net{} achieves state-of-the-art performance.

The main contributions of this work are as follows:
\begin{enumerate}[topsep=0pt]
    \item We propose the novel \net{} for radar localization in prior LiDAR maps, addressing both place recognition and metric localization tasks.
    \item We propose to solve the metric localization task by predicting pixel-level matches in the form of a flow field between the radar and the LiDAR \ac{BEV} images.
    \item We extensively evaluate \net{} against state-of-the-art place recognition and metric localization methods on three real-world datasets.
    \item We investigate the generalization ability of our method by evaluating it in a different city and using a different sensor setup than the ones used for training.
    \item We release the code and trained models at \url{http://ralf.cs.uni-freiburg.de}.
\end{enumerate}

\begin{figure*}[!t]
    \centering
    \includegraphics[width=0.68\linewidth]{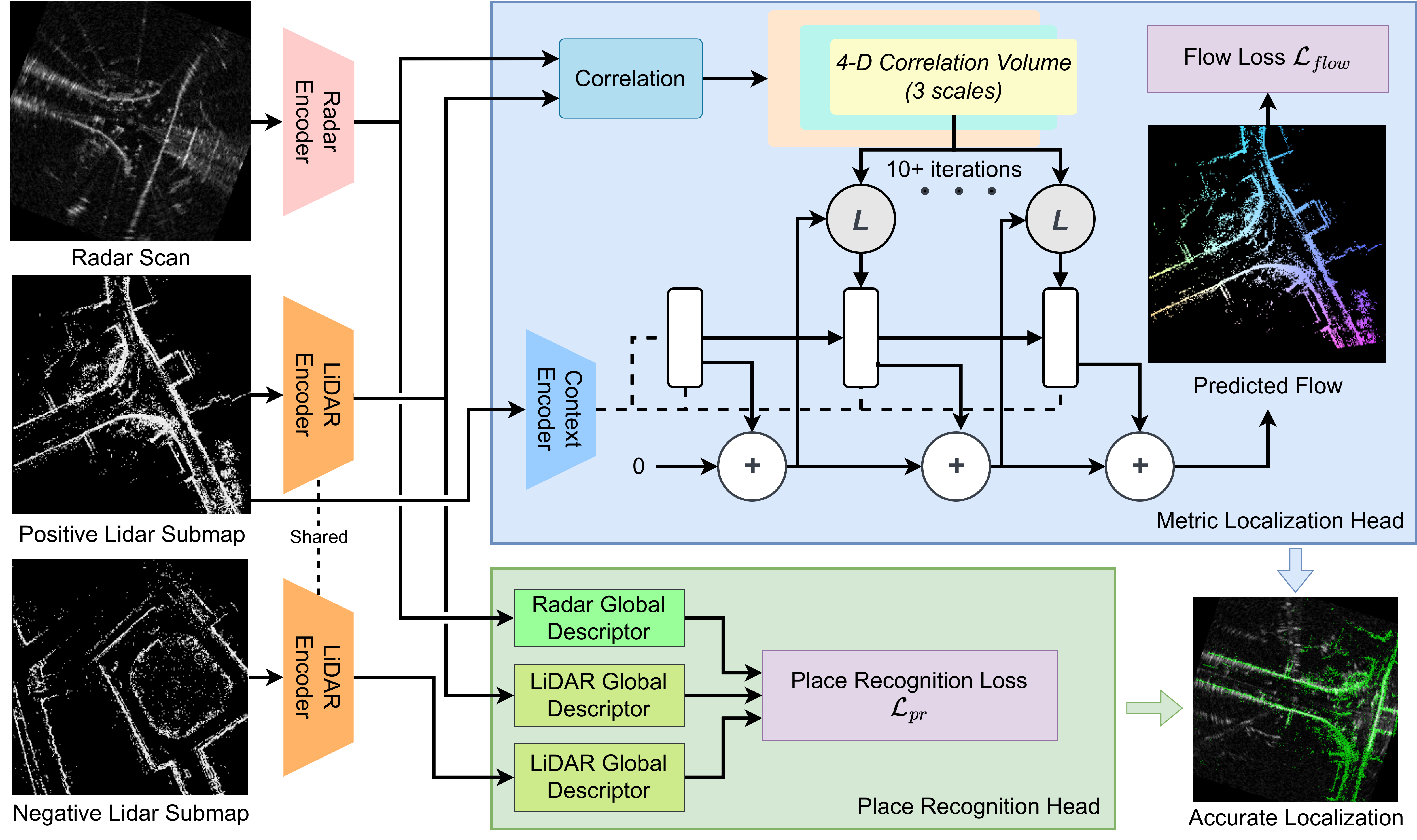}
    \caption{Overview of \net{} for joint place recognition and metric localization of radar scans in a LiDAR map. It consists of feature encoders, a place recognition head to extract global descriptors, and a metric localization head to estimate the 3-DoF pose of the query radar scan within the LiDAR map.}
    \label{fig:method}
    \vspace{-15pt}
\end{figure*}

\section{Related Work}

In this section, we discuss related work on LiDAR/radar place recognition and metric localization, for both the same-modality and cross-modality settings.

{\parskip=3pt
\noindent\textit{Place Recognition}:
Place recognition has been extensively explored over recent decades. 
Scan Context~\cite{kim2018scan} and M2DP~\cite{he2016m2dp} investigate place recognition with LiDAR data using hand-crafted descriptors.
With the advent of \acp{CNN}, DiSCO~\cite{xu2021disco} employs a learning-based approach for LiDAR-based place recognition using Scan Context as input. 
On the other hand, Gadd~\textit{et~al.}~\cite{gadd2021contrastive} tackle radar-based place recognition through unsupervised contrastive loss learning. They additionally highlight the importance of carefully selecting positive and negative samples to facilitate training the network with a contrastive loss function.
Additionally, Cait~\textit{et~al.}~\cite{cait2022autoplace} leverage data from a single-chip automotive radar (point cloud) for place recognition. 
While these works primarily focus on single sensor modalities, there have been notable efforts involving cross-modal data.\looseness=-1

In the context of cross-modal place recognition, addressing the disparities between diverse sensor modalities requires the establishment of a cohesive embedding space.
Various strategies have emerged to achieve this goal.
Recent research efforts employ shared networks to create joint embeddings, unifying data from different modalities~\cite{barsan2020learning, cattaneo2020global}.
Conversely, Yin~\textit{et~al.}~\cite{yin2023radar} suggest leveraging \acp{GAN} to transform data from one modality into a newly generated sample that resembles the other modality.
Furthermore, Radar-to-lidar~\cite{yin2021radar} employs shared network processing of radar and LiDAR Scan Contexts to generate feature embeddings, subsequently employing KD-Trees for clustering. 
In this application, a shared encoder is employed under the assumption that Scan Contexts for both radar and LiDAR modalities exist within the same embedding space.\looseness=-1 

Our approach distinguishes itself by employing \ac{BEV} images of radar and LiDAR modalities for place recognition.
Our solution hinges on a novel \ac{CNN}-based method that crafts joint-space embeddings between the two modalities. 
Notably, the use of \ac{BEV} images facilitates metric localization as well, a feat unattainable through alternative data representations such as Scan Context, which only provides an estimation of the angle between two samples.

{\parskip=3pt
\noindent\textit{Metric Localization} pertains to precisely estimating the position of a robot within a map of the environment. Historically, metric localization was addressed with classical robotics techniques involving probabilistic updates to adjust the likelihood of a robot's location on a map~\cite{fox2003bayesian,fox2001particle,chen2011kalman}. Recent progress has enabled learning-based techniques to achieve accurate metric localization.
LCDNet~\cite{cattaneo2022lcdnet} detects loop closures in LiDAR point clouds and estimates relative scan-to-map poses. 
On the other hand, OverlapNet~\cite{chen2021overlapnet} gauges the overlap between LiDAR scans and uses ICP-based techniques for relative pose estimation.}

Past endeavors have seen the application of cross-modal strategies in metric localization.
CMRNet~\cite{cattaneo2019cmrnet,cattaneo2020cmrnetpp} employs \ac{CNN}-based techniques to localize a camera image onto a pre-existing LiDAR map. 
In the method proposed by Tang~\textit{et~al.}~\cite{tang2021get}, LiDAR scans are localized against aerial satellite images. 
To manage the distinct modalities, an encoder-decoder network creates an occupancy map from the overhead images. 
These maps are subsequently transformed into point clouds and registered against LiDAR scans using point-based methods.
Tang~\textit{et~al.}~\cite{tang2021self} propose to estimate relative poses in a self-supervised manner by identifying the augmented sample with optimal rotation and translation noise that aligns most closely with the current overhead image.
RaLL~\cite{yin2021rall} proposes the use of a differentiable measurement model to localize radar samples on a pre-existing LiDAR map.
This measurement model is then applied to a Kalman filter, thus making the whole learning system differentiable. In contrast to previous approaches, we tackle the metric localization challenge as a flow estimation task. 
We compute flow vectors between radar and LiDAR \ac{BEV} images to establish initial pixel correspondences.
Subsequently, we employ \ac{RANSAC}~\cite{fischler1981random} to estimate the accurate relative transformations between the two inputs.

\section{Technical Approach}

In this section, we describe \net{}, our proposed approach for place recognition and metric radar localization in LiDAR maps. An overview of \net{} is illustrated in \figref{fig:method}. The architecture of our approach is built upon RAFT~\cite{teed2020raft}, a state-of-the-art network for optical flow estimation. \net{} comprises three main components: feature extraction, place recognition head, and metric localization head. In the rest of this section, we detail each of the components and the respective loss functions, followed by a description of the inference procedure.\looseness=-1

\subsection{Feature Extraction}
\label{sec:encoders}
The architecture of the two encoders, namely the radar encoder and LiDAR encoder, is based on the feature encoder of RAFT~\cite{teed2020raft}, which consists of a convolutional layer with stride equal to two, followed by six residual layers with downsampling after the second and fourth layer. Differently from the original feature encoder of RAFT which shares weights between the two input images, \net{} employs separate feature extractors for each modality due to the distinct nature of radar and LiDAR data.
Formally, given a radar \ac{BEV} image $\mathbf{R} \in \mathbb{R}^{H \times W \times 1}$ and a LiDAR \ac{BEV} image $\mathbf{L} \in \mathbb{R}^{H \times W \times 1}$, the two encoders $g_r$ and $g_l$ extract features at one-eight of the original resolution $g_r, g_l: \mathbb{R}^{H \times W \times 1} \rightarrow \mathbb{R}^{H/8 \times W/8 \times D}$. The features extracted by the two encoders are shared between the place recognition head and the metric localization head.

\subsection{Place Recognition Head}

The place recognition head has a twofold purpose: firstly, it aggregates the feature maps from the feature extractor into a global descriptor. Secondly, it maps features from radar and LiDAR data, which naturally lie in different embedding spaces, into a shared embedding space, where global descriptors of radar scans and LiDAR submaps can be compared against each other.
The architecture of the place recognition head is a shallow \ac{CNN} composed of four convolutional layers with feature sizes (256, 128, 128, 128), respectively. Each convolutional layer is followed by batch normalization and ReLU activation. Differently from the feature encoders, the place recognition head is shared between the radar and LiDAR modalities.

To train the place recognition head, we use the well-known triplet technique~\cite{schroff2015facenet}, where triplets composed of (anchor, positive, negative) samples are selected to compute the triplet loss. The positive sample is a \ac{BEV} image depicting the same place as the anchor sample, while the negative sample is a \ac{BEV} image of a different place. While typically this technique is employed to compare triplets of samples of the same modality, in our case the samples can be generated from different modalities. For instance, given an anchor radar scan $\mathbf{R}^a$, a positive LiDAR submap $\mathbf{L}^p$, and a negative LiDAR submap $\mathbf{L}^n$, we define the triplet loss $\mathcal{L}_{tr}^{RLL}$ as
\begin{equation}
    \label{eq:loss_triplet}
    \mathcal{L}_{tr}^{RLL} = \max \left\{ d( \mathbf{F}_R^\text{a}, \mathbf{F}_L^\text{p} ) - d( \mathbf{F}_R^\text{a}, \mathbf{F}_L^\text{n} ) + m , 0 \right\},
\end{equation}
where $\mathbf{F}_R^a$, $\mathbf{F}_L^p$, and $\mathbf{F}_L^n$, are the global descriptors of $\mathbf{R}^a$, $\mathbf{L}^p$, $\mathbf{L}^n$, respectively; $m$ is the triplet margin, and $d(\cdot)$ is a given distance function. The superscript $RLL$ of $\mathcal{L}_{tr}^{RLL}$ represent the modalities of the (anchor, positive, negative) samples, in this case (radar, LiDAR, LiDAR). We apply the same loss to all the eight possible combination of modalities, leading to the final place recognition loss:
\begin{equation}
  \begin{split}
    \label{eq:loss_triplet_final}
    \mathcal{L}_{pr} = & \mathcal{L}_{tr}^{RRR} + \mathcal{L}_{tr}^{RLL} + \mathcal{L}_{tr}^{RLR} + \mathcal{L}_{tr}^{RRL} +\\
    &  \mathcal{L}_{tr}^{LLL} + \mathcal{L}_{tr}^{LRR} + \mathcal{L}_{tr}^{LRL} + \mathcal{L}_{tr}^{LLR}.
  \end{split}
\end{equation}

To select the triplets that compose a batch, we first randomly sample a positive sample for each anchor sample. We define a sample to be positive of an anchor sample if the distance between their position is smaller than a positive threshold $\tau_p$. Furthermore, we select the hardest negative sample from the batch of samples currently being processed by the network, making sure that its position is farther away from the anchor than a negative threshold $\tau_n$. This technique is known as online hardest negative mining.

\begin{figure*}[t]
    \centering  
    \subfloat[\label{fig:oxford}Oxford Radar Robotcar]{\includegraphics[width=0.25\textwidth]{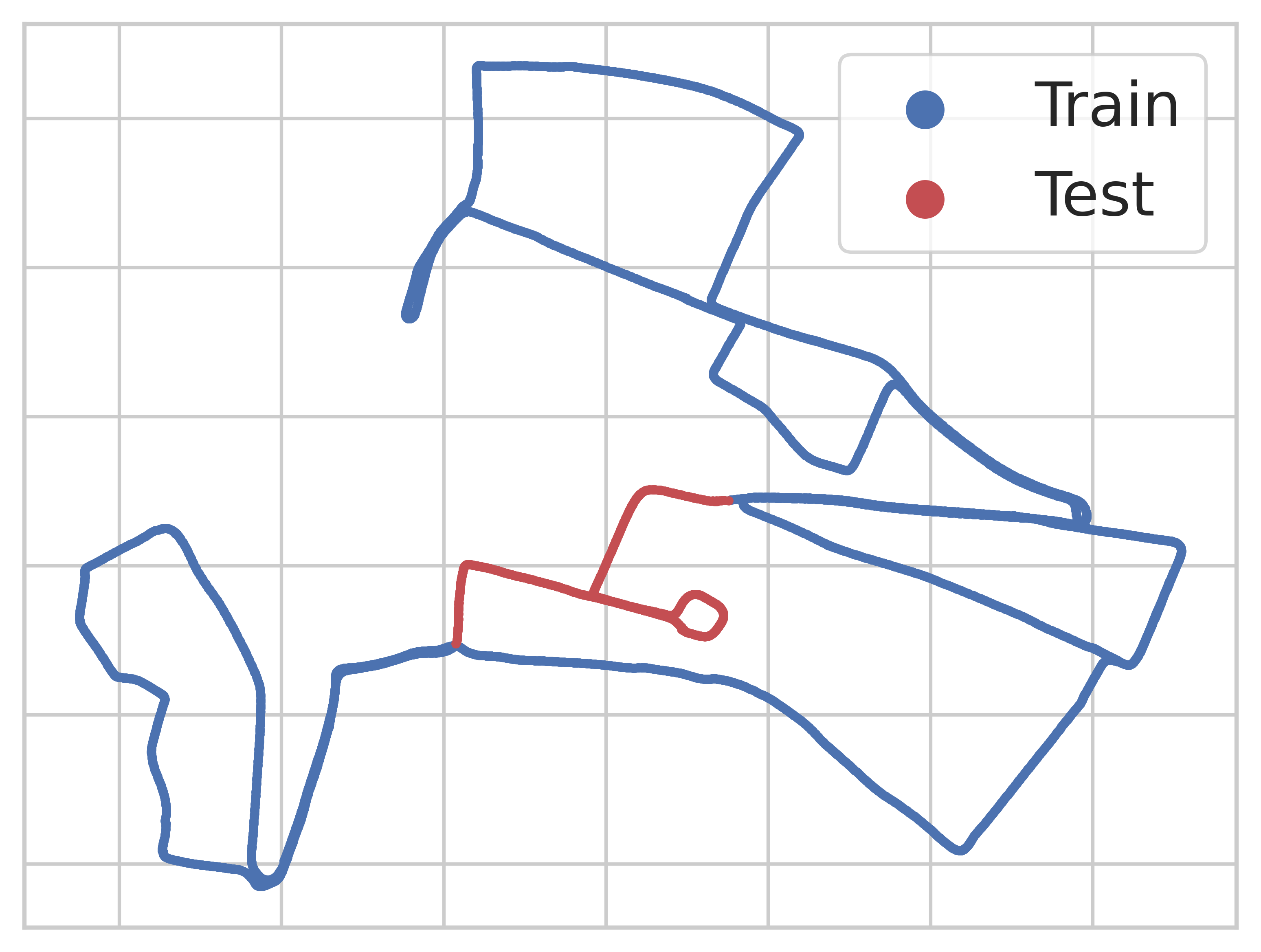}}
    \hspace{2em}
    \subfloat[\label{fig:kaist}MulRan-Kaist]{\includegraphics[width=0.25\textwidth]{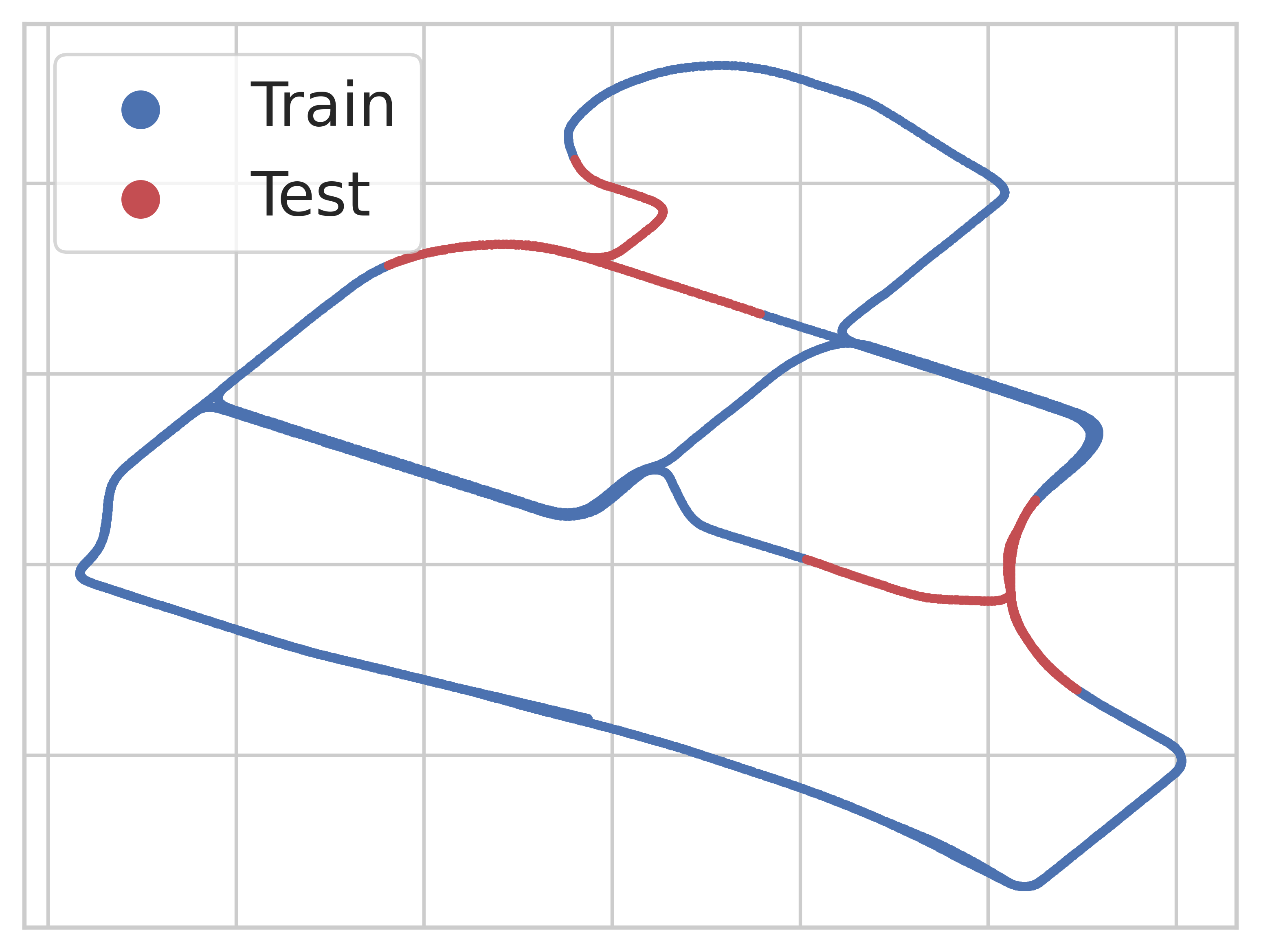}}
    \hspace{2em}
    \subfloat[\label{fig:boreas}Boreas]{\includegraphics[width=0.25\textwidth]{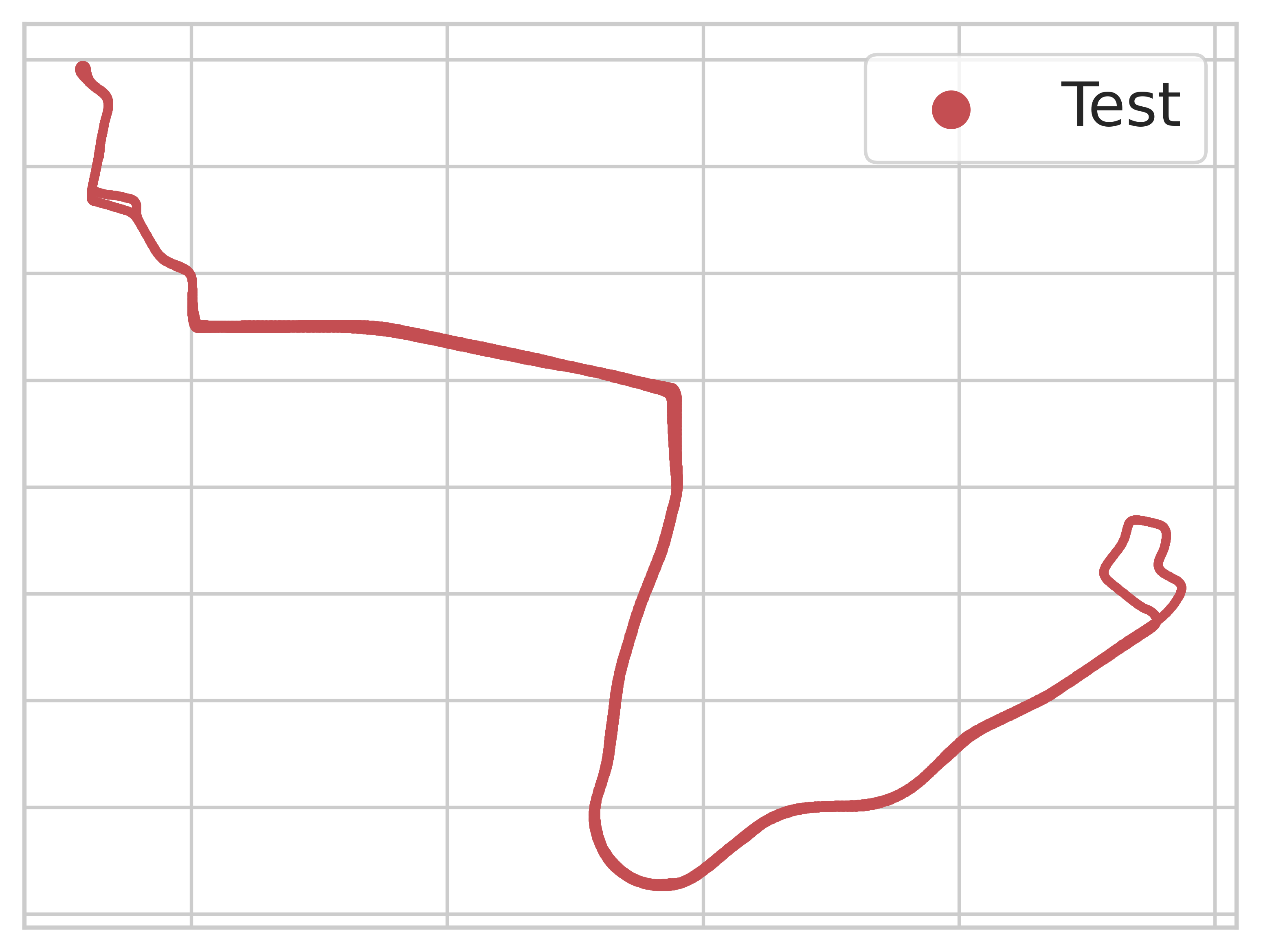}}
    \caption{Train-test split of the three datasets used in our experiments. The blue and red trajectories represent the train and test splits, respectively.}
    \label{fig:datasets}
  \vspace{-13pt}
\end{figure*}

\subsection{Metric Localization Head}
For metric localization of radar scans against a LiDAR map $\mathcal{M} $, we propose to learn pixel-wise matches in the form of flow vectors. The intuition behind this decision is that a radar \ac{BEV} image and a LiDAR \ac{BEV} image taken at the same position should be well aligned, as depicted in the bottom right part of \figref{fig:method}. Therefore, for every pixel in the LiDAR \ac{BEV} image, our metric localization head predicts the corresponding pixel in the radar \ac{BEV} image.

More formally, given a radar \ac{BEV} image $\mathbf{R}$, and the initial coarse pose $T_{init}$ predicted by the place recognition head, we generate a LiDAR \ac{BEV} image $\mathbf{L}$ centered around $T_{init}$. The metric localization head takes the two \ac{BEV} images $\mathbf{R}$ and $\mathbf{L}$ as input, and predicts a dense flow map $\mathbf{f}$ that aligns the two images. Each pixel $(u, v)$ in $\mathbf{f}$ contains the flow vector $(\Delta u, \Delta v)$ that maps the pixel $\mathbf{L}_{(u, v)}$ to the pixel $\mathbf{R}_{(u + \Delta u, v + \Delta v)}$.
The architecture of our metric localization head is based on RAFT~\cite{teed2020raft}, which first computes a 4-D correlation volume between the features extracted by the two encoders as described in \secref{sec:encoders}. The correlation volume is then fed into a \ac{GRU} that iteratively refines the estimated flow map. Each iteration $i$ of the \ac{GRU} update outputs a flow update $\Delta \mathbf{f}_i$, which is added to the previous flow estimate $\mathbf{f}_{i-1}$ to obtain the updated flow map $\mathbf{f}_i$. Following~\cite{teed2020raft}, we employ an additional context encoder that extracts features only from the LiDAR \ac{BEV} image, which is additionally fed to the \ac{GRU}.
To generate the ground truth flow map $\mathbf{f}_{GT}$, we first transform the LiDAR map points in the initial pose $T_{init}$, and compute their pixel position in the relative \ac{BEV} image $\mathbf{L}$ as follows:
\begin{equation}
    [ u_{init}, v_{init} ] = f_{bev}(T_{init} \cdot \mathcal{M}),
\end{equation}
where $f_{bev}$ is the function that projects a 3D point cloud into the \ac{BEV} image.
Similarly, we compute the pixel position of the projection when transforming the map using the ground truth pose $T_{GT}$:
\begin{equation}
    [ u_{GT}, v_{GT} ] = f_{bev}(T_{GT} \cdot \mathcal{M}).
\end{equation}
Finally, we compute the ground truth flow map $\mathbf{f}_{GT}$ by comparing the projected points using the initial pose and the ground truth pose as
\begin{equation}
    \mathbf{f}_{GT} = [ u_{GT} - u_{init}, v_{GT} - v_{init} ].
\end{equation}

To train the metric localization head, we use the loss function originally proposed in RAFT~\cite{teed2020raft}, which supervises the predicted flow maps $\mathbf{f}_i$ at each iteration of the \ac{GRU} as
\begin{equation}
    \label{eq:loss_flow}
    \mathcal{L}_{flow} = \sum_{i = 1}^N \gamma^{N - i}||\mathbf{f}_{GT} - \mathbf{f}_i||_1,
\end{equation}
where $\gamma=0.8$ gives exponentially increasing weights to later iterations. Due to the sparse nature of the \ac{BEV} images, we only compute the loss on the non-zero pixels in $\mathbf{L}$.

The final loss function of \net{} is the sum of the individual loss functions defined in \eqref{eq:loss_triplet_final} and \eqref{eq:loss_flow}:
\begin{equation}
    \mathcal{L}_{total} = \mathcal{L}_{pr} + \mathcal{L}_{flow}.
\end{equation}

\subsection{Inference}
\label{sec:inference}

Before deployment, we split the LiDAR map $\mathcal{M}$ into multiple overlapping submaps $\mathcal{M}_i$, with the relative poses $T_i$, and we generate a LiDAR \ac{BEV} image $\mathbf{L}_i$ for each submap. We then generate the global descriptors $\mathbf{F}_L^i$ for each submap $\mathbf{L}_i$ using the LiDAR encoder and the place recognition head.

During inference, given a query radar scan $\mathbf{R}$, we first compute its global descriptor $\mathbf{F}_R$, and we compare it against all the submap descriptors $\mathbf{L}_i$. We then select the submap $\mathbf{L}_k$ with the highest similarity to $\mathbf{R}$:
\begin{equation}
    k = \text{argmin}_i \|  \mathbf{F}_R - \mathbf{F}_L^i \|_2 \ .
\end{equation}

We then feed $\mathbf{F}_R$ and $\mathbf{L}_k$ to the metric localization head to predict the flow map $\mathbf{f} = [\mathbf{f}^u, \mathbf{f}^v]$, which we use to generate a warped LiDAR \ac{BEV} image $\mathbf{L}_k^{warp}$ as
\begin{equation}
    \mathbf{L}_k^{warp} (u + \mathbf{f}^u_{(u,v)}, v + \mathbf{f}^v_{(u,v)}) = \mathbf{L}_k (u, v).
\end{equation}
Subsequently, we convert the two images $\mathbf{L}_k$ and $\mathbf{L}_k^{warp}$ into point clouds $\mathbf{P}_k$ and $\mathbf{P}_k^{warp}$, by multiplying each pixel location by the pixel resolution of the \ac{BEV} image, and setting the height to zero. Finally, we use \ac{RANSAC} to estimate the query radar pose $T_{pred}$ that minimizes the distance between the two point clouds:\looseness=-1
\begin{equation}
    T_{pred} = \text{argmin}_{T} \| \mathbf{P}_k^{warp} - T \cdot \mathbf{P}_k \|_2 \ .
\end{equation}

\begin{table*}[t]
  \centering
  \caption{Comparison with the state of the art in terms of recall@1 ($\SI{3}{\meter}$).}
  \label{table:recall_r2r_l2l_r2l}
  \begin{tabular}{l|ccc|ccc|ccc}
  \toprule
  \multirow{2}{*}{Approach} & & Oxford~\cite{barnes2020oxford} & & & MulRan-Kaist~\cite{kim2020mulran} & & & Boreas~\cite{burnett2023boreas} &  \\
  \cmidrule{2-10}
  & R2R & {L2L} & {R2L} & {R2R} & {L2L} & {R2L} & {R2R} & {L2L} & {R2L} \\
  \midrule
  M2DP~\cite{he2016m2dp}                      & - & 0.20 & - & - & 0.34 & - & - & 0.74 & - \\
  Scan Context~\cite{kim2018scan}              & 0.87 &0.97 & 0.016 & \textbf{0.97} & 0.97 & 0.02 & 0.96 & 0.89 & 0.002 \\
  RaPlace~\cite{jang2023raplace}              & 0.78 & - & - & 0.83 & - & - & 0.87  & - & - \\
  DiSCO~\cite{xu2021disco}                    & 0.90 & 0.96 & 0.013 & \textbf{0.97} & \textbf{0.98} & 0.05 & 0.96 & 0.93 & 0.001 \\
  Radar-to-LiDAR~\cite{yin2021radar}          & 0.85 & 0.93 & 0.56 & 0.90 & 0.89 & 0.46 & 0.96 & 0.92 & 0.05 \\
  \net{} (Ours)                               & \textbf{0.97} & \textbf{0.98} & \textbf{0.63} & 0.88 & 0.89 & \textbf{0.58} & \textbf{0.99} & \textbf{0.99} & \textbf{0.71} \\
  \bottomrule
  \end{tabular}
  \vspace{-10pt}
\end{table*}

\section{Experimental Evaluation}
In this section, we first describe the datasets that we use for training and evaluation, followed by details on our training protocol. We then present results from evaluating \net{} against the state-of-the-art in both place recognition and metric localization. Finally, we perform multiple ablation studies to validate the design choices of our method.

\subsection{Datasets}
We evaluate our proposed approach on three real-world driving datasets, namely Oxford Radar Robotcar~\cite{barnes2020oxford}, MulRan~\cite{kim2020mulran} and Boreas~\cite{burnett2023boreas}. We use the Oxford and the MulRan datasets for training and evaluation, while we use the Boreas dataset for testing the generalization ability of \net{} in a different city and a different sensor setup than those used for training.
For the Robotcar dataset, we use the train-test split proposed in~\cite{yin2023radar}, while for MulRan, we use the KAIST sequences and selected two random geographical areas as the test set. \figref{fig:datasets} shows the train-test split of the three datasets\footnote{On Oxford, we used the sequences 2019-01-18-12-42-34, 2019-01-18-14-14-42, 2019-01-18-14-46-59, and 2019-01-18-15-20-12 for training. For testing, we used the sequence 2019-01-18-15-20-12 as a map and the sequence 2019-01-18-14-46-59 as a query. On KAIST, we used sequences 02 and 03 for training, and for testing, sequence 02 was used as a query against sequence 03. On Boreas, the sequence 2021-05-06-13-19 serves as a query while the sequence 2021-05-13-16-11 as the map.}.
It is important to note that we do not train a separate model for each dataset, instead, we train a single model on the combined training split of the Oxford and MulRan datasets and evaluate it on the test splits of all three datasets.

\subsection{Training Details}
We use the PyTorch deep learning library for model training and evaluation on a machine equipped with an Intel i5-6500@3.2GHz processor and two NVIDIA TITAN-X GPUs. We use images of size 256 $\times$ 256 at a resolution of \SI{0.5}{\meter}/pixel for the \ac{BEV} projection.
For data augmentation, we apply random rotations and translations in the range $\pm 30^{\circ}$ and $\pm \SI{5}{\meter}$, respectively, for both radar and LiDAR samples.
We use the AdamW optimizer and the OneCycle learning rate scheduler with a learning rate of $5\cdot10^{-4}$.
The duration of the learning rate increase is 10\% of the whole training time, which is about $2\cdot10^{5}$ iterations.
We train the network with a batch size of 15 (anchor, positive) pairs per iteration.
During training, we set $m=0.5$, $\tau_p = \SI{2}{\meter}$ and $\tau_n = \SI{80}{\meter}$, and the distance function $d(\cdot)$ as the L2 distance.

\begin{figure}[t]
  \centering  
  \subfloat[Oxford Radar Robotcar]{\includegraphics[width=0.5\linewidth]{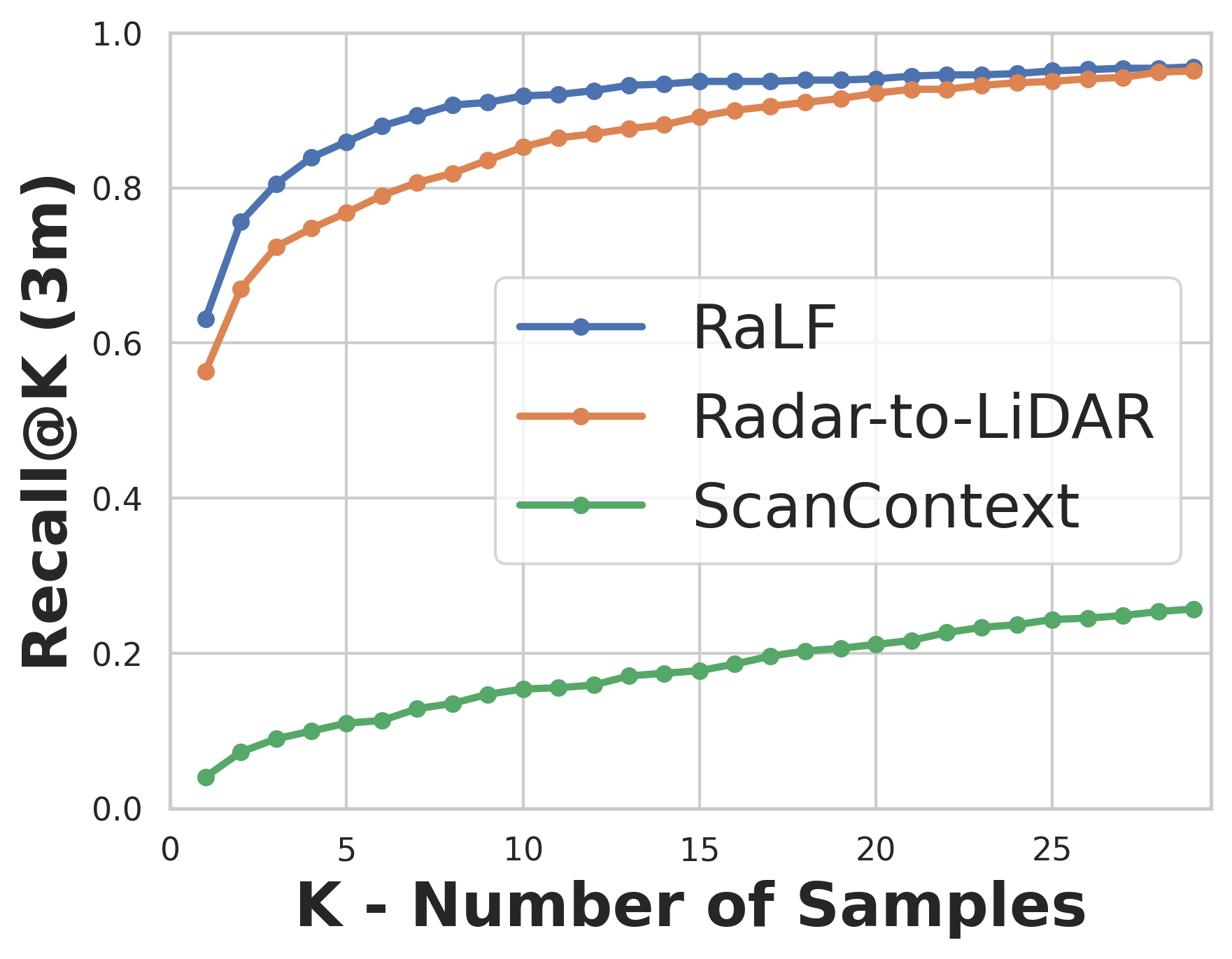}}
  \hfill
  \subfloat[MulRan-Kaist]{\includegraphics[width=0.5\linewidth]{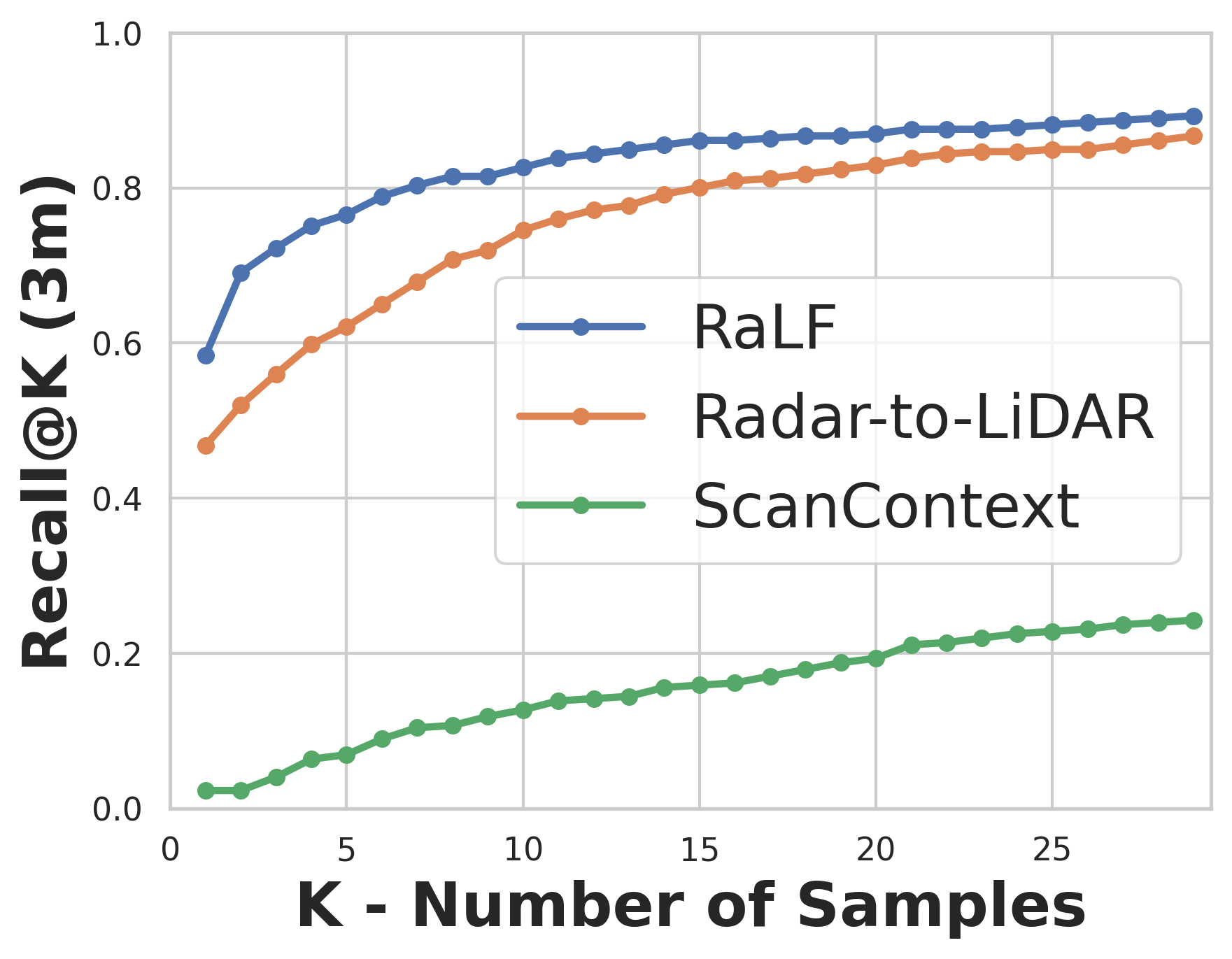}}
  \hfill
  \subfloat[Boreas]{\includegraphics[width=0.5\linewidth]{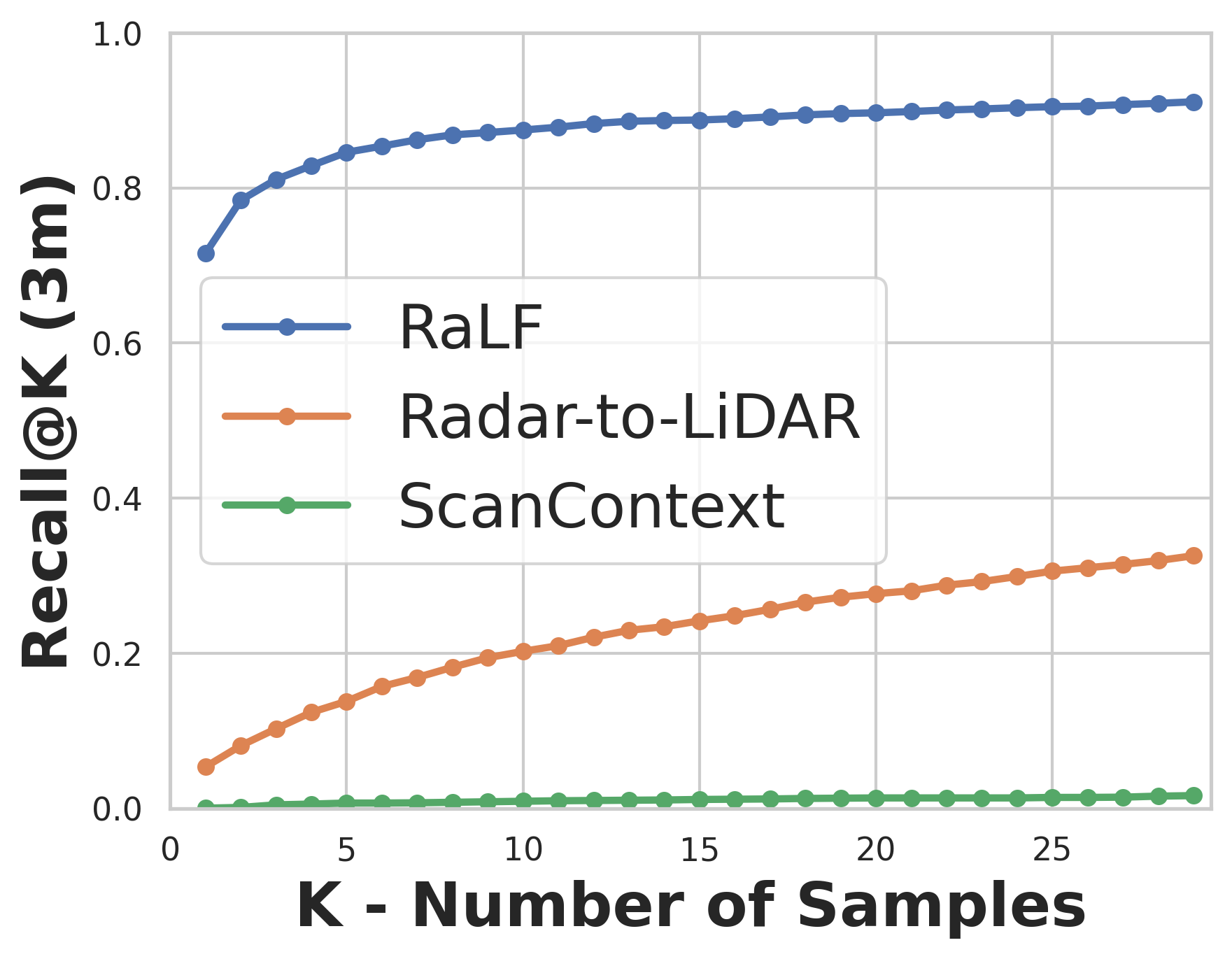}}
  \caption{Recall@k ($\SI{3}{\meter}$) at different values of k. }
  \label{fig:recallatk}
\vspace{-10pt}
\end{figure}

\subsection{Place Recognition}

We compare the place recognition performance of \net{} against handcrafted methods M2DP~\cite{he2016m2dp}, Scan Context~\cite{kim2018scan}, and RaPlace~\cite{jang2023raplace} as well as learning-based methods DiSCO~\cite{xu2021disco} and Radar-to-lidar~\cite{yin2021radar}. For all the above methods, we use the implementation provided by the respective authors. To provide a fair comparison, we retrained all learning-based methods on the same training set used to train \net{}. Radar-to-lidar is the only existing method that specifically tackles the cross-modal radar-LiDAR place recognition task. RaPlace, on the other hand, focuses on radar-radar place recognition. All the other methods were originally proposed for LiDAR place recognition, although some of them can easily be adapted to radar data. We use the recall@1 (3m) metric, which considers a query as correctly localized if the pose of the most similar database sample is within three meters from the pose of the query.\looseness=-1

\tabref{table:recall_r2r_l2l_r2l} summarizes the results from this experiment. We observe that our method outperforms all the other methods in the radar-LiDAR place recognition task. In particular, we achieve a recall@1 of 0.63, 0.58, and 0.71 on the Oxford, MulRan, and Boreas datasets, respectively. This is a significant improvement over the state-of-the-art method Radar-to-lidar~\cite{yin2021radar}, which achieves a recall@1 of 0.56, 0.46, and 0.05 on the same datasets. We also observe that \net{} achieves state-of-the-art performance in the radar-radar and LiDAR-LiDAR place recognition tasks on both the Oxford and Boreas datasets. \figref{fig:recallatk} shows the recall@k ($\SI{3}{\meter}$) at different values of k for all the datasets. We observe that our method outperforms the state-of-the-art at all values of k, especially at smaller values of k.

\subsection{Metric Localization}

\begin{table}[t]
\centering
\caption{Comparison of relative pose errors (rotation and translation) between positive pairs on the Oxford Radar Robotcar dataset.} 
\label{table:rot_trans_errors}
\begin{tabular}{l|ccc}
\toprule
{Approach} & {$\delta x $} (m) {$\downarrow$} & {$\delta y$} (m) {$\downarrow$} & {$\delta \theta$} (Deg) {$\downarrow$} \\
\midrule
RaLL~\cite{yin2021rall}  & 1.04 & \textbf{0.69} & 1.26 \\
\midrule
\net{} (Ours), (RaLL params) & \textbf{0.97} & 0.96 & \textbf{1.15} \\
\net{} (Ours), ($\pm 30^\circ, \pm \SI{5}{\meter}$) & 1.07 & 1.03 & 1.26 \\
\bottomrule
\end{tabular}
\vspace{-13pt}
\end{table}

In \tabref{table:rot_trans_errors}, we report the rotation and translation errors between the estimated transformation and the ground truth transformation.
The results show that our method outperforms RaLL~\cite{yin2021rall} while using the same data augmentation scheme ($\pm 6^{\circ}, \pm \SI{6}{\meter}$) during evaluation.
We achieve a mean rotation error of $1.26^{\circ}$ and translation errors of $\SI{1.07}{\meter}$ and $\SI{1.03}{\meter}$ along X and Y-directions respectively. In \figref{fig:qualitative_flow}, we report the qualitative results of metric localization, where the query radar scans are overlaid with LiDAR submaps using the initial coarse pose $T_{init}$, the pose estimated by \net{} $T_{pred}$, and the ground truth pose $T_{GT}$.\looseness=-1

\begin{figure}[t]
  \captionsetup[subfigure]{labelformat=empty}
  \centering  
  \subfloat{\includegraphics[width=0.31\linewidth]{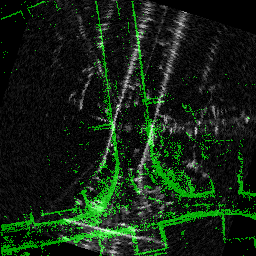}}
  \hfill
  \subfloat{\includegraphics[width=0.31\linewidth]{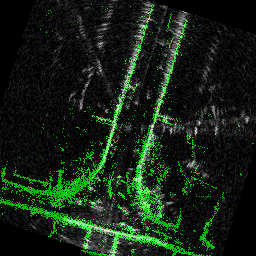}}
  \hfill
  \subfloat{\includegraphics[width=0.31\linewidth]{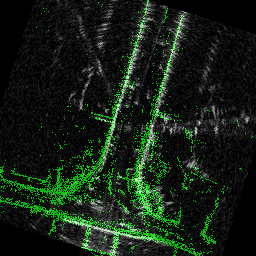}}

  \vspace{5pt}

  \subfloat{\includegraphics[width=0.31\linewidth]{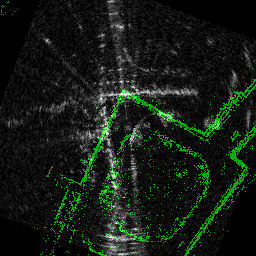}}
  \hfill
  \subfloat{\includegraphics[width=0.31\linewidth]{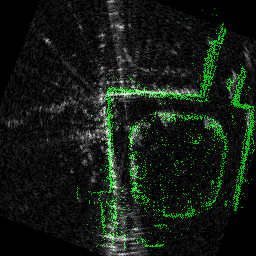}}
  \hfill
  \subfloat{\includegraphics[width=0.31\linewidth]{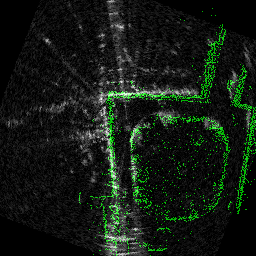}}

  \vspace{5pt}

  \subfloat[Initial Pose]{\includegraphics[width=0.31\linewidth]{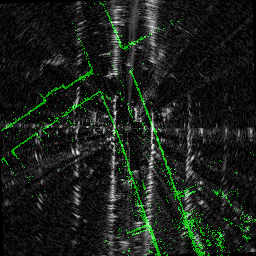}}
  \hfill
  \subfloat[\net{} alignment]{\includegraphics[width=0.31\linewidth]{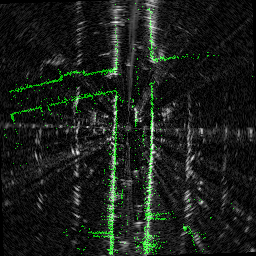}}
  \hfill
  \subfloat[GT alignment]{\includegraphics[width=0.31\linewidth]{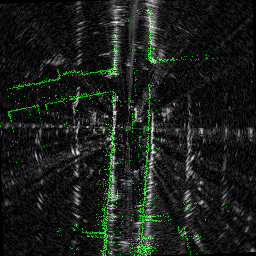}}

  \caption{Qualitative results of radar scans (grayscale) aligned with the LiDAR submaps (green) using our proposed method.}
  \label{fig:qualitative_flow}
\end{figure}

\subsection{Ablation Studies}\looseness=-1

\begin{table}[t]
\centering
\caption{Ablation study on the encoder architecture.}
\begin{tabular}{l|c|cccc}
\toprule
{Encoder} & {$\#$} Params & {Recall@1} & {$\delta x$} & {$\delta y$} & {$\delta \theta$} \\
& {(M)} & {($\SI{3}{\meter}$)} {$\uparrow$} & {($\SI{}{\meter}$)} {$\downarrow$} & {($\SI{}{\meter}$)} {$\downarrow$} & {(Deg)} {$\downarrow$} \\
\midrule
VGG-19 & 21 & 0.54 & 1.10 & 1.11 & 1.49 \\
ResNet-18 & 6.4 & 0.52 & 1.28 & 1.25 & 1.60 \\
ResNet-34 & 7.8 & 0.52 & 1.17 & 1.03 & 1.54 \\
ResNet-50 & 8.9 & 0.60 & 1.26 & 1.04 & 1.42 \\
RAFT-C & 6.4 & 0.04 & 5.71 & 3.28 & 14.5 \\
RAFT-S & 2.0 & 0.66 & 1.32 & 1.30 & 1.51 \\
\textbf{RAFT} & 7.4 & \textbf{0.67} & \textbf{1.07} & \textbf{1.03} & \textbf{1.26} \\
\bottomrule
\end{tabular}
\vspace{-10pt}
\label{table:encoder_choice}
\end{table}

We perform multiple ablation studies to validate the design choices that we made while developing \net{}.
All the experiments reported in this section are trained and tested on the Oxford Radar RobotCar dataset.

\subsubsection*{Encoder Choice}
In the feature extraction module, we employ different encoder architectures, including ResNet~\cite{he2016deep}, VGGNet~\cite{simonyan2014very}, and RAFT encoders~\cite{teed2020raft}, to process \ac{BEV} images and generate feature embeddings. 
\tabref{table:encoder_choice} summarizes the results of this study which shows that the RAFT encoder performs best for our approach.
The smaller RAFT-S version also achieves similar recall@1 with just 2 Million parameters.
Furthermore, when we used a common encoder for both the radar and LiDAR images in RAFT-C (Common), the network was not able to learn a meaningful representation for the two heads, emphasizing the need to process different sensor modalities with separate encoders.\looseness=-1

\subsubsection*{Choice of Place Recognition Head}
We performed experiments with various \ac{CNN} architectures, each with a different number of convolutional layers, for the place recognition head. 
Additionally, we evaluated the NetVLAD layer, which has shown success in \ac{VPR} tasks.
The results of this experiment are presented in \tabref{table:pr_head_choice}. Interestingly, we observed that NetVLAD underperforms in comparison to \ac{CNN}-based heads in our setting. 
We hypothesize that this can be attributed to the sparsity of features extracted from the \ac{BEV} images and the cross-modal nature of feature comparison. 
Furthermore, the \ac{CNN} architectures show increasingly better performance while increasing the number of layers up to four.
\begin{table}[t]
\centering
\caption{Ablation study on the place recognition architecture.}
\begin{tabular}{l|c}
\toprule
{PR Head} & {Recall@1 ($\SI{3}{\meter}$)} {$\uparrow$} \\
\midrule
NetVLAD & 0.12 \\
CNN-2 & 0.57 \\
CNN-3 & 0.62 \\
\textbf{CNN-4} & \textbf{0.67} \\
CNN-5 & 0.64 \\
\bottomrule
\end{tabular}
\vspace{-10pt}
\label{table:pr_head_choice}
\end{table}

\begin{table}
\centering
\caption{Ablation study on the place recognition loss function.}
\begin{tabular}{l|c}
\toprule
{$\mathcal{L}_{pr}$} & {Recall@1 ($\SI{3}{\meter}$)} {$\uparrow$} \\
\midrule
Contrastive Loss & 0.19 \\
NPairs Loss & 0.26 \\
Quadruplet Loss & 0.43 \\
Triplet Loss - Cosine & 0.58 \\
Triplet Loss - L1 & 0.51 \\
\textbf{Triplet Loss - L2} & \textbf{0.67} \\
\bottomrule
\end{tabular}
\vspace{-10pt}
\label{table:metric_loss_func}
\end{table}

\subsubsection*{Choice of Metric Loss Function}

Finally, we experiment with different loss functions for the place recognition head.
We evaluate the contrastive loss, N-pairs loss, triplet margin loss, and quadruplet loss.
For the triplet loss, we additionally evaluate different distance functions $d(\cdot)$ in~\eqref{eq:loss_triplet}: Cosine Similarity, L1 and L2.
As this change only affects the place recognition head, we exclusively report the recall@1 scores for this experiment.
\tabref{table:metric_loss_func} shows the results from this experiment. We observe that the triplet margin loss outperforms all the other losses, while the quadruplet loss is the next best loss function for our use case.
This is probably due to the hard-negative mining strategy that we use while mining for triplets. We also observe that the L2 distance function achieves the best performance.

\section{Conclusion}
In this paper, we proposed \net{}, a novel approach to simultaneously address place recognition and metric localization of radar scans in LiDAR maps.
To the best of our knowledge, \net{} is the first approach that combines both these tasks in an end-to-end manner.
Our method is composed of feature encoders, a place recognition head, and a metric localization head based on flow vectors.
We presented extensive experiments on the Oxford Radar Robotcar and MulRan datasets, and demonstrated that it achieves state-of-the-art performance for both place recognition and metric localization.
Furthermore, we demonstrated that our method can effectively generalize to different cities and sensor setups than the ones used during training by evaluating it on the Boreas dataset. To foster future research, we made the code for \net{} and the trained models publicly available.

\footnotesize
\bibliographystyle{IEEEtran}
\balance
\bibliography{references.bib}

\end{document}